\documentclass{article}
\usepackage{spconf,amsmath,graphicx,hyperref}
\usepackage[acronym]{glossaries}
\usepackage{booktabs}
\usepackage{algorithm}
\usepackage[noend]{algpseudocode}
\usepackage[table]{xcolor}
\usepackage{balance}
\usepackage{tikz}
\usepackage{stfloats}

\definecolor{TMIBlue}{RGB}{0, 138, 218}
\definecolor{IEEEBlue}{RGB}{0, 67, 147}

\title{Nuclear Diffusion Models for Low-Rank\\Background Suppression in Videos}

\name{
    Tristan S.W. Stevens$^{\star}$, \quad
    Oisín Nolan$^{\star}$, \quad Jean-Luc Robert$^\dagger$\quad
    Ruud J.G. van Sloun$^{\star}$
    }

\address{$^{\star}$Dept. of Electrical Engineering, Eindhoven University of Technology, the Netherlands\\$^{\dagger}$Philips Research North America, Cambridge MA, USA}

\usepackage{amsmath,amsfonts,bm,mathtools}

\def\1{\bm{1}}

\def\rvepsilon{{\mathbf{\epsilon}}}

\def\rvn{{\mathbf{n}}}

\def\rvx{{\mathbf{x}}}
\def\rvy{{\mathbf{y}}}

\def\rmI{{\mathbf{I}}}

\def\mI{{\bm{I}}}

\def\mL{{\bm{L}}}

\def\mX{{\bm{X}}}
\def\mY{{\bm{Y}}}

\DeclareMathAlphabet{\mathsfit}{\encodingdefault}{\sfdefault}{m}{sl}
\SetMathAlphabet{\mathsfit}{bold}{\encodingdefault}{\sfdefault}{bx}{n}

\newcommand{\R}{\mathbb{R}}

\newcommand{\norm}[1]{\left \lVert #1 \right \rVert}

\newcommand{\defeq}{\vcentcolon=}

\newacronym{awgn}{AWGN}{additive white Gaussian noise}
\newacronym{dgm}{DGM}{deep generative model}
\newacronym{dps}{DPS}{diffusion posterior sampling}
\newacronym{dm}{DM}{diffusion model}
\newacronym{lpips}{LPIPS}{learned perceptual image patch similarity}
\newacronym{psnr}{PSNR}{peak signal-to-noise ratio}
\newacronym{ef}{EF}{ejection fraction}
\newacronym{snr}{SNR}{signal-to-noise ratio}
\newacronym{mi}{MI}{mechanical index}
\newacronym{gan}{GAN}{generative adversarial network}
\newacronym{vae}{VAE}{variational autoencoder}
\newacronym{mmse}{MMSE}{minimum mean square error estimator}
\newacronym{ema}{EMA}{exponential moving average}
\newacronym{seqdiff}{SeqDiff}{sequential diffusion}
\newacronym{cs}{CS}{compressed sensing}
\newacronym{rf}{RF}{radio-frequency}
\newacronym{ood}{OoD}{out-of-distribution}
\newacronym{cdm}{CDM}{conditional diffusion model}
\newacronym{pca}{PCA}{principal component analysis}
\newacronym{ica}{ICA}{independent component analysis}
\newacronym{rpca}{RPCA}{robust PCA}
\newacronym{pcp}{PCP}{principal component pursuit}
\newacronym{gcnr}{gCNR}{generalized contrast-to-noise ratio}
\newacronym{ks}{KS}{Kolmogorov–Smirnov}

\begin{document}
\maketitle
\begin{abstract}
Video sequences often contain structured noise and background artifacts that obscure dynamic content, posing challenges for accurate analysis and restoration. Robust principal component methods address this by decomposing data into low-rank and sparse components. Still, the sparsity assumption often fails to capture the rich variability present in real video data. To overcome this limitation, a hybrid framework that integrates low-rank temporal modeling with diffusion posterior sampling is proposed. The proposed method, \textit{Nuclear Diffusion}, is evaluated on a real-world medical imaging problem, namely cardiac ultrasound dehazing, and demonstrates improved dehazing performance compared to traditional RPCA concerning contrast enhancement (gCNR) and signal preservation (KS statistic). These results highlight the potential of combining model-based temporal models with deep generative priors for high-fidelity video restoration.
\end{abstract}
\vspace{-0.1cm}
\section{Introduction}
\vspace{-0.1cm}
\label{sec:intro}
Denoising, the recovery of a clean signal from a corrupted observation, is a foundational problem in signal processing~\cite{milanfar2024denoising}, encompassing a diverse range of applications from natural image and video enhancement to sensory applications such as medical imaging and radar~\cite{stevensDeepGenerative2025}. Typically, the objective is to disentangle informative structure from nuisance variability, thereby improving interpretability and downstream analysis of observed data. In sequential data, such as videos, the corruption is often structured rather than random, with background artifacts and stationary patterns that can obscure the dynamic content of interest.

\Acrfull{rpca} provides a principled approach for separating foreground dynamics from structured background by decomposing sequential data into a \emph{low-rank} background and a \emph{sparse} component, representing the signal of interest. This idea has been widely applied in various domains, including speech~\cite{mirbeygi2021rpca}, saliency detection, face recognition, and medical imaging~\cite{bouwmans2018applications}. While sparsity can be imposed in alternative transform domains (e.g., wavelet or frequency bases), such choices remain handcrafted and often insufficient to capture the rich variability of real signals.

Meanwhile, \acrfullpl{dgm} are well-suited to capturing complex statistical dependencies, such as those present in videos. \acrfullpl{dm}~\cite{ho2020denoising,song2020score} in particular, have opened new directions in image restoration. \Acrshortpl{dm} learn to sample from complex data distributions through iterative denoising, and have achieved state-of-the-art performance across inverse problems such as denoising, deblurring, and dehazing~\cite{daras2024survey}. Acting as expressive learned priors, \Acrshortpl{dm} can capture rich statistical structure beyond simple pixel-wise sparsity priors.

Building on this idea, we propose a hybrid denoising framework that combines the temporal modeling of \acrshort{rpca} with the expressive modeling of diffusion priors. In the proposed method, the conventional sparsity assumption on the foreground component is replaced with a learned diffusion prior, while maintaining a nuclear norm penalty to encourage low-rank temporal structure of the background. This is realized through \acrfull{dps}~\cite{chung2022diffusion}, which alternates between reverse diffusion and measurement-guided updates, ultimately allowing more accurate recovery of the dynamic foreground.

While our formulation is general-purpose, we evaluate our method on the real-world problem of cardiac ultrasound video dehazing, where structured noise (haze) degrades image quality and hampers diagnostic clarity~\cite{stevens2024dehazing}. \Acrshort{rpca} has been a popular tool in ultrasound imaging, for example in clutter suppression~\cite{solomon2019deep} and elastography denoising~\cite{ashikuzzaman2020denoising}, and has been extended through deep unfolding for high-dimensional settings~\cite{cai2021learned}. Despite these advances, the sparsity assumption is often too restrictive in practice.\\

In this paper, we make the following contributions:
\begin{itemize}
  \item A novel framework integrating \acrshort{dps} with a low-rank temporal model, generalizing \acrshort{rpca} via data-driven deep generative priors.
  \item An evaluation of the method on cardiac ultrasound dehazing, achieving enhanced image contrast while better preserving anatomical structures compared to standard \acrshort{rpca}.
\end{itemize}
\begingroup
\renewcommand\thefootnote{}
\footnotetext{\scriptsize\url{https://tue-bmd.github.io/nuclear-diffusion/}}
\endgroup
\clearpage

\section{Background}
\label{sec:background}

\subsection{Robust PCA for background supression}
\label{sec:rpca}

\Acrfull{rpca} decomposes observations  $\mY\in\R^{n\times p}$ (e.g., pixel intensities of $p$ frames, each of size $n$) into:
\begin{equation}
\mY \;=\; \mL + \mX,
\label{eq:pca}
\end{equation}
where $\mL$ is low-rank (coherent background, e.g., static haze) and $\mX$ is sparse (foreground dynamics, e.g. tissue signal). Exact rank minimization is intractable, so \acrshort{rpca} solves the convex surrogate known as \acrfull{pcp}. A common approach is to relax the hard equality constraint and form the Lagrangian:
\begin{equation}
\min_{\mL,\mX}\; \|\mL\|_* + \lambda\,\|\mX\|_1
+ \frac{\mu}{2}\,\|\mY - \mL - \mX\|_F^2,
\label{eq:rpca-lang}
\end{equation}
where $\norm{\cdot}_F$ denotes the Frobenius norm, and $\|\cdot\|_*$ denotes the nuclear norm, which serves as a convex surrogate for the rank of $\mL$. In many imaging scenarios, the assumed signal model of \eqref{eq:pca} is overly simplistic. The signal component $\mX$ often exhibits complex, structured patterns that are not truly sparse. As a result, standard \acrshort{rpca} tends to over-penalize these patterns, effectively attenuating or removing portions of the true signal. This limitation motivates replacing the generic $\ell_1$ sparsity prior with a learned diffusion prior that better models the complex distribution of images, while retaining low-rank temporal modeling through a nuclear norm constraint.

\begin{figure}
    \centering
    \includegraphics[width=1\linewidth]{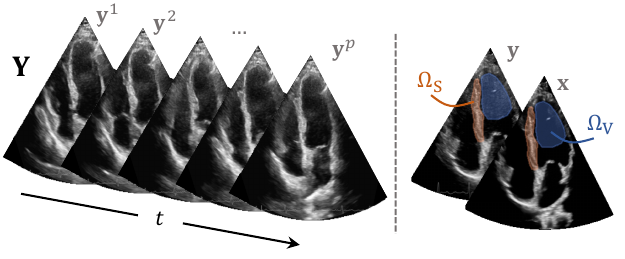}
    \vspace{-0.8cm}
    \caption{Sequence of hazy cardiac ultrasound images $\mY$, with annotated regions of interest $\Omega$ used for evaluation.}
    \label{fig:sequence}
\end{figure}
\subsection{Diffusion Posterior Sampling}
\label{sec:dps}

Diffusion models learn the distribution $p(\rvx)$ of a random variable $\rvx$ through a forward corruption process that gradually adds Gaussian noise:
\begin{equation}
    \rvx_\tau = \alpha_\tau \rvx_0 + \sigma_\tau \rvepsilon, \quad \rvepsilon \sim \mathcal{N}(\mathbf{0}, \rmI),
    \label{eq:forward-diffusion}
\end{equation}
where \(\rvx_0 \equiv \rvx \sim p(\rvx)\), \(\tau \in [0, \mathcal{T}]\), and \(\alpha_\tau, \sigma_\tau\) are predefined noise schedules. The generative process is then defined as reversing this corruption, which is equivalent to iterative denoising of \(\rvx_\tau\), starting from ${\rvx_{\mathcal{T}} \sim \mathcal{N}(\mathbf{0}, \rmI)}$. Tweedie's formula relates the minimum mean-square error estimate to the score of the distribution:
\begin{equation}
    \rvx_{0 \mid \tau} \defeq\, \mathbb{E}[\rvx_0\vert\rvx_\tau] = \frac{1}{\alpha_\tau} \left( \rvx_\tau + \sigma_\tau^2 \nabla_{\rvx_\tau} \log p(\rvx_\tau) \right),
    \label{eq:tweedie}
\end{equation}
where \(\rvx_{0 \mid \tau}\) is a one-step denoised estimate, and the score \(\nabla_{\rvx_\tau} \log p(\rvx_\tau)\) is parameterized by a neural network $\rvepsilon_\theta$, often predicting the noise, which relates to the score via $\rvepsilon_\theta(\rvx_\tau, \tau)\approx - \sigma_\tau\nabla_{\rvx_\tau} \log p(\rvx_\tau)$. To sample from the prior distribution $p(\rvx)$, at each step $\tau$, $\rvx_0$ is estimated using \eqref{eq:tweedie} and mapped back to $\rvx_{\tau-1}$ via forward diffusion in \eqref{eq:forward-diffusion}, ensuring a smooth sampling trajectory. The score network is trained using the denoising score matching objective~\cite{vincent_connection_2011}.
\emph{Unconditional} score-based diffusion models can be adapted for \emph{conditional} sampling given noisy measurements, i.e., generating $\rvx \sim p(\rvx \mid \rvy)$, by computing the score of the Bayesian posterior. Exact posterior sampling with diffusion models is generally intractable, but several approximate methods exist~\cite{daras2024survey}. Here, we adopt \acrfull{dps}~\cite{chung2022diffusion}, which interleaves prior updates (denoising) with guidance steps (gradient steps toward the measurements). The prior update follows the same procedure as in unconditional sampling, while the guidance step, with forward model $\rvy = f(\rvx) + \rvn$ and $\rvn\sim\mathcal{N}(0, \sigma_{\rvn}^2\mI)$ is given by:
\begin{align}
\nabla_{\rvx_\tau} \log p(\rvy \mid \rvx_\tau) &\approx \nabla_{\rvx_\tau} \log p\big(\rvy \mid \rvx_{0\mid\tau} \big) \label{eq:dps-linear-1} \\
&= -\frac{1}{2\sigma_{\rvn}^2} \nabla_{\rvx_\tau} \norm{\rvy - f(\rvx_{0\mid\tau})}_2^2. \label{eq:dps-linear-2}
\end{align}
We note that while $p(\rvy \mid \rvx_0)$ is often known exactly, the noise-perturbed likelihood $p(\rvy \mid \rvx_\tau)$ generally does not admit a closed form, motivating the approximation in \eqref{eq:dps-linear-1}.

\begin{table}
  \makebox[\linewidth][l]{\hspace{-0.4cm}%
    \begin{minipage}{\dimexpr\linewidth+0.4cm\relax}
      \caption{Comparison of RPCA and Nuclear Diffusion.}
      \label{tab:priors}
      \begin{tabular}{lcc}
        \toprule[1.2pt]
         & \textbf{RPCA} & \textbf{Nuclear Diffusion} \\
        \midrule
        $p(\mY \mid \mL, \mX)$ & $\mathcal{N}(\mY; \mL+\mX, \sigma^2 \mathbf{I})$ & $\mathcal{N}(\mY; \mL+\mX, \sigma^2 \mathbf{I})$ \\
        $p(\mL)$ & $\propto \exp(-\|\mL\|_*)$ & $\propto \exp(-\|\mL\|_*)$ \\
        $p(\mX)$ & $\propto \exp(-\lambda \|\mX\|_1)$ & \textcolor{purple}{$p_\theta(\mX)$} \\
        \arrayrulecolor{darkgray}
        \addlinespace[0.5ex]
        \hline
        \addlinespace[0.5ex]
        \arrayrulecolor{black}
        Inference & \hspace{-0.4cm}$\underset{\mL,\mX}{\arg\max}\; p(\mL,\mX \mid \mY)$ & \textcolor{purple}{$\mX, \mL \sim p_\theta(\mX,\mL \mid \mY)$} \\
        \bottomrule[1.2pt]
      \end{tabular}
    \end{minipage}
  }
\end{table}

\section{Methods}
\label{sec:methods}
We adopt a Bayesian perspective to generalize the \acrshort{rpca} framework and extend it with a learned diffusion prior. Given observations $\mY \in \R^{n \times p}$ and independent latent variables $\mL$ and $\mX$ we construct the following joint distribution:
\begin{equation}
    p(\mY, \mL, \mX) = p(\mY \mid \mL, \mX)\,p(\mL)\,p(\mX).
    \label{eq:bayes}
\end{equation}
To arrive at the RPCA objective in \eqref{eq:rpca-lang}, one can use a Gaussian forward model for the likelihood term:
\begin{equation}
    p(\mY \mid \mL, \mX) = \mathcal{N}(\mY; \mL+\mX, \mu^{-1} \mathbf{I}).
    \label{eq:rpca-likelihood}
\end{equation}
Similarly, the low-rank component $\mL$ follows a nuclear norm prior:
\begin{equation}
    p(\mL) \propto \exp(-\gamma \|\mL\|_*),
    \label{eq:prior-L}
\end{equation}
which can be interpreted as a low-rank inducing prior, while the signal component $\mX$ is modeled with a Laplace prior to enforce sparsity:
\begin{equation}
    p(\mX) \propto \exp(-\lambda \|\mX\|_1).
    \label{eq:prior-X-laplace}
\end{equation}
Taking the negative logarithm of \eqref{eq:bayes} and selecting the point estimate that maximizes the posterior (i.e., the MAP solution) recovers the classical RPCA objective in Lagrangian form, as shown in \eqref{eq:rpca-lang}.

\algnewcommand{\RequireOptional}{\item[\textbf{Optional:}]}
\algrenewcommand\algorithmiccomment[1]{\hfill\textcolor{gray!60}{$\triangleright$~#1}}
\newcommand{\CommentLine}[1]{%
  \Statex \hspace*{\ALG@thistlm} \textcolor{gray!60}{$\triangleright$~#1}%
}

\begin{algorithm}
\caption{Nuclear Diffusion Posterior Sampling}
\begin{algorithmic}[1]
\Require observations $\mY$, low-rank weighting $\gamma$, guidance weighting $\mu$, diffusion model $\rvepsilon_\theta$, diffusion steps $\mathcal{T}$, noise schedule $\alpha_\tau, \sigma_\tau$
\State Initialize $\mX_\mathcal{T} \sim \mathcal{N}(\mathbf{0}, \sigma_\mathcal{T}^2 \mathbf{I})$, $\mL \gets \mathbf{0}$
\For{$\tau = \mathcal{T}$ to $0$}

    \State $\rvepsilon^t \gets \rvepsilon_\theta(\rvx^t_\tau, \tau), \; \forall t=1,\dots,p$ \Comment{Predict noise}
    \State  \textcolor{purple}{$\mathbf{E}_\tau \gets [\rvepsilon^1, \dots, \rvepsilon^p]$} \Comment{Stack frames}
    \State $\mX_{0|\tau} \gets \frac{1}{\alpha_\tau} (\mX_\tau - \sigma_\tau \mathbf{E}_\tau)$ \Comment{Denoise (prior)}
    
    \State $\mathcal{E}_\tau \gets \frac{\mu}{2}\|\mY - \mL - \mX_{0|\tau}\|_F^2$ \Comment{Measurement error}
    \State $\mX_{0|\tau} \gets \mX_{0|\tau} - \nabla_{\mX} 
    \mathcal{E}_\tau$ \Comment{Likelihood guidance}
    \State $\mX_{\tau-1} \gets \alpha_\tau \mX_{0|\tau} + \sigma_\tau \rvepsilon$ \Comment{Forward diffusion}
    \State \textcolor{purple}{$\mathcal{R}_\tau \gets \gamma \|\mL\|_*$} \Comment{Low-rank penalty}
    \State \textcolor{purple}{$\mL \gets \mL - \nabla_{\mL} (\mathcal{E}_\tau +\mathcal{R}_\tau)$} \Comment{Background update}
\EndFor
\State \Return $\mX_0, \mL$
\end{algorithmic}
\label{algo:robust-diff}
\end{algorithm}
\vspace{-0.5cm}

\subsection{Nuclear diffusion}
Building on this probabilistic formulation, we propose a hybrid framework that replaces the $\ell_1$ sparsity prior on $\mX$ with a learned diffusion prior $p_\theta(\mX)$ and performs posterior sampling instead of a MAP estimate, i.e., $\mX, \mL \sim p_\theta(\mX, \mL \mid \mY)$. This allows $\mX$ to capture complex, structured patterns beyond simple sparsity, while retaining the low-rank nuclear norm prior on $\mL$ for temporal coherence. In practice, we implement this by interleaving a reverse diffusion process, as described in Section~\ref{sec:dps}, with gradient-based guidance from the likelihood in \eqref{eq:rpca-likelihood} and the low-rank prior.

The diffusion prior $p_\theta(\mX)$ is applied in the spatial domain to individual frames, while temporal dependencies are enforced solely through the low-rank prior on $\mL$. This separation enables the use of pretrained 2D diffusion models, simplifying implementation and avoiding the need for specialized video diffusion networks. Concretely, the signal component $\mX$ can be written as:
\begin{equation}
    \mX = \begin{bmatrix} \rvx^{1} & \rvx^{2} & \dots & \rvx^{p} \end{bmatrix}\in \R^{n\times p},
\end{equation}
where each column $\rvx^t \in \R^n$ represents the vectorized image at time $t$. The learned prior is parameterized by a denoising diffusion model $\rvepsilon_\theta(\rvx^t_\tau, \tau)$, which independently operates on a single noisy frame $\rvx^t_\tau$ at diffusion step $\tau$:
\begin{equation}
    \rvx^t_\tau \mapsto \rvepsilon_\theta(\rvx^t_\tau, \tau),
    \quad \forall \, t \in \{1,\dots,p\}.
\end{equation}
Applying the per-frame denoiser independently across all frames induces a joint prior over the entire sequence, formalized in terms of scores for diffusion sampling~\cite{song2020score}:
\begin{equation}
\nabla_{\mX} \log p_\theta(\mX) = -\frac{1}{\sigma_\tau} \;\big[\rvepsilon_\theta(\rvx_\tau^1, \tau), \dots, \rvepsilon_\theta(\rvx_\tau^p, \tau)\big].
\end{equation}
The inference framework is detailed in Algorithm~\ref{algo:robust-diff} and a comparison of the distributions is given in Table~\ref{tab:priors}.
\begin{figure}
    \centering
    \includegraphics[width=0.9\linewidth]{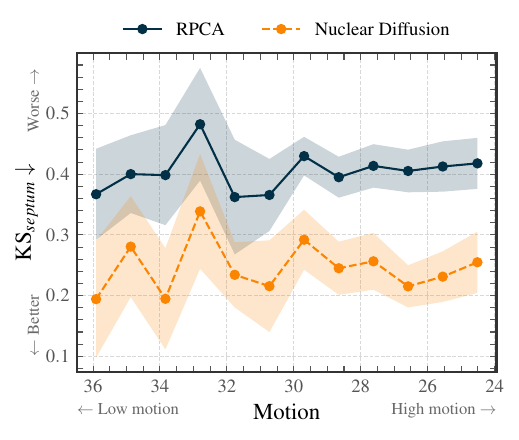}
    \vspace{-0.3cm}
    \caption{\acrshort{ks} statistic for various amount of motion levels measured via $\text{PSNR}(\rvy^t, \rvy^{t-1})$, with Nuclear Diffusion outperforming RPCA across the entire range.}
    \label{fig:motion}
\end{figure}
\section{Results}
\label{sec:results}
\begin{figure*}
    \centering
    \includegraphics[width=1\linewidth]{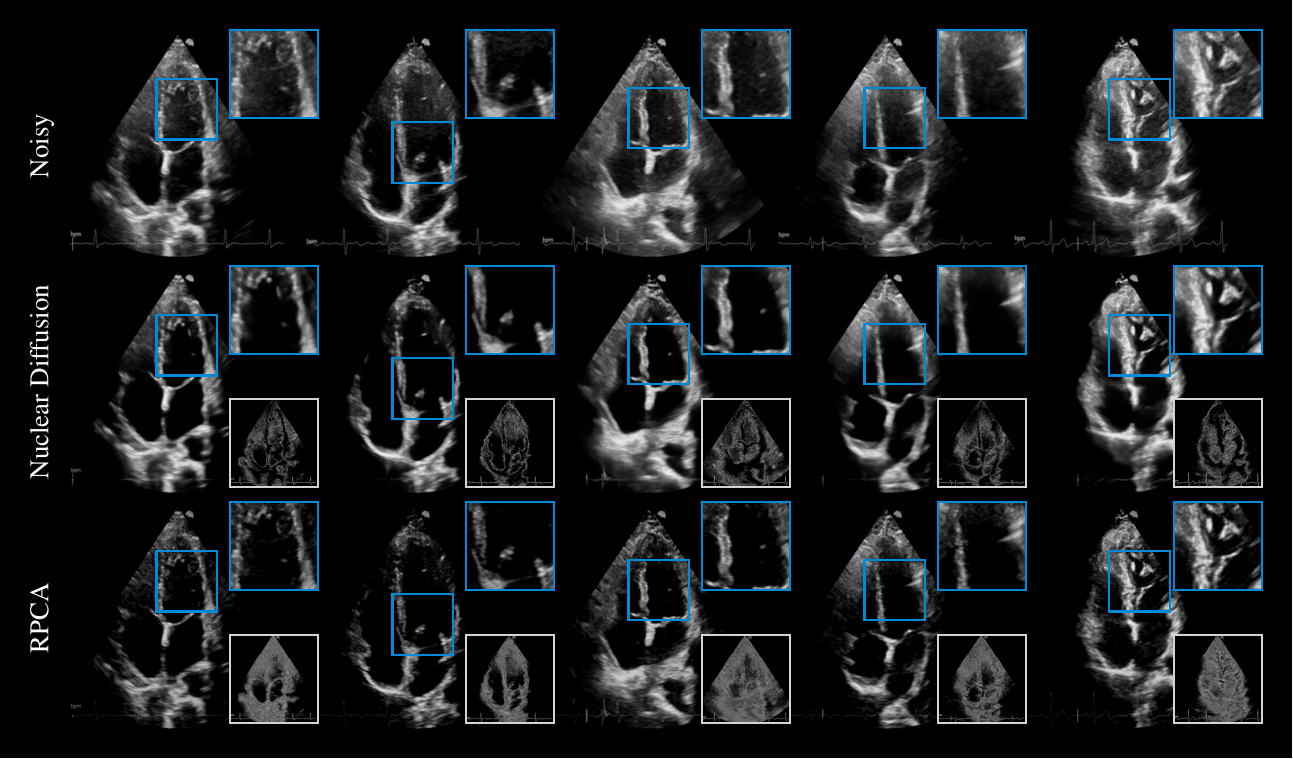}
    \vspace{-0.4cm}
    \caption{Comparison on the task of  cardiac ultrasound dehazing. While both methods suppress haze (shown in bottom insets), \acrshort{rpca} tends to excessively attenuate tissue, resulting in sparse structures, whereas Nuclear Diffusion better preserves details.}
    \vspace{-0.2cm}
    \label{fig:qualitative-comparison}
\end{figure*}

\begin{figure}
    \centering
    \includegraphics[width=0.95\linewidth]{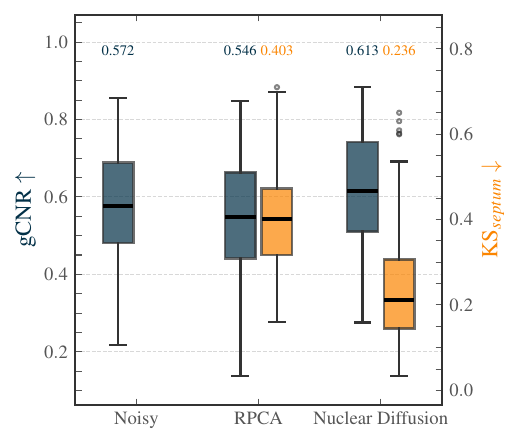}
    \caption{Quantitative comparison of Nuclear Diffusion and \acrshort{rpca} using \acrshort{gcnr} and \acrshort{ks} metrics.
    Nuclear Diffusion achieves higher contrast between $\Omega_S$ and $\Omega_V$ while preserving the tissue intensity distribution in $\Omega_S$, whereas \acrshort{rpca} tends to attenuate tissue and distort signal statistics.}
    \label{fig:quantitative}
\end{figure}

We evaluate the proposed method on the task of cardiac ultrasound dehazing, focusing on both haze removal and tissue structure preservation. Given that a ground truth is not available, performance is assessed using two unsupervised metrics: \acrfull{gcnr}~\cite{rodriguez2019generalized}, which we use to measure contrast between ventricle $\Omega_V$ and septum $\Omega_S$ regions.
Additionally, we use the \acrfull{ks} statistic to quantify agreement between the original $\mY$ and denoised $\mX$ tissue distributions in the septum region  $\Omega_S$.
\begin{equation}
    \text{KS} = \sup_z \Big| F_{\Omega_S(x)}(z) - F_{\Omega_S(y)}(z) \Big|,
\end{equation}
where $F(\cdot)$ is the empirical CDF of the respective ROIs.
See Fig.~\ref{fig:sequence} for an example sequence with annotated regions. The dataset contains videos of $60\times256\times256$, with in total 4,376 clean frames from 75 easy-to-image subjects and 2,324 noisy frames from 40 difficult-to-image subjects~\cite{DehazingEcho2025}. Fig.~\ref{fig:qualitative-comparison} presents a qualitative comparison. Both Nuclear Diffusion and RPCA reduce haze, but RPCA over-attenuates tissue, resulting in sparse or discontinuous structures, while Nuclear Diffusion adheres better to the prior trained on the clean dataset. Quantitative results support these observations, as shown in Fig.~\ref{fig:quantitative} and Fig.~\ref{fig:motion}. Results are generated with 500 diffusion steps, accelerated using SeqDiff~\cite{stevens2024sequential} ($\mathcal{T}=5000$), with $\mY^t$ as initialization. Furthermore, we use $\gamma=1$, $\mu=2$, and $p=7$. The method is implemented using the zea~\cite{zea2025} library with JAX backend.

\vspace{-0.1cm}
\section{Conclusions}
In this paper, we introduced a hybrid framework that generalizes \acrshort{rpca} by integrating low-rank temporal modeling with learned generative diffusion priors. By replacing the standard $\ell_1$ sparsity prior with a score-based generative model and performing diffusion posterior sampling with a nuclear norm penalty, our approach captures complex signal components while explicitly separating dynamic foreground from low-rank background. We demonstrated the effectiveness of this method on cardiac ultrasound video dehazing, showing that it can suppress haze artifacts and improve image contrast while preserving delicate tissue . These results highlight the potential of combining classical low-rank priors with modern generative models for video restoration.
\vspace{-1cm}

\clearpage
\bibliographystyle{IEEEbib}
\balance
\bibliography{references}

\end{document}